\documentclass{article}
\pdfoutput=1
\usepackage[nonatbib,final]{nips_2016}

\usepackage[utf8]{inputenc} 
\usepackage[T1]{fontenc}    
\usepackage{hyperref}       
\usepackage{url}            
\usepackage{booktabs}       
\usepackage{amsfonts}       
\usepackage{nicefrac}       
\usepackage{microtype}      

\usepackage{graphicx}
\usepackage{amsmath}
\DeclareMathOperator*{\argmax}{argmax}

\title{Full-Time Supervision based Bidirectional RNN for Factoid Question Answering}

\author{
  Dong Xu,~ Wu-Jun Li \\
  National Key Laboratory for Novel Software Technology \\
  Collaborative Innovation Center of Novel Software Technology and Industrialization\\
  Department of Computer Science and Technology, Nanjing University, China \\
  \texttt{121220312@smail.nju.edu.cn, liwujun@nju.edu.cn} \\
}

\begin{document}

\maketitle

\begin{abstract}
  Recently, bidirectional recurrent neural network~(BRNN) has been widely used for question answering~(QA) tasks with promising performance. However, most existing BRNN models extract the information of questions and answers by directly using a pooling operation to generate the representation for loss or similarity calculation. Hence, these existing models don't put supervision~(loss or similarity calculation) at every time step, which will lose some useful information. In this paper, we propose a novel BRNN model called full-time supervision based \mbox{BRNN}~(FTS-BRNN), which can put supervision at every time step. Experiments on the factoid QA task show that our FTS-BRNN can outperform other baselines to achieve the state-of-the-art accuracy.

\end{abstract}

\section{Introduction}
Question answering~(QA) has become an important research topic in natural language processing~(NLP) with wide applications. Factoid QA~(e.g., quiz bowl) is a special QA task, which has also attracted much attention recently~\cite{Boyd-Graber:12,Iyyer:14}. Traditional QA methods can be divided into three main categories. The first category is based on surface pattern matching which uses manually defined rules~\cite{Riloff:00} or parse dependency trees~\cite{Wang:07,Cui:05}. This category of methods need a lot of human work and usually need to be modified when the data is changed. The second category is based on similarity measures which usually uses bag-of-words~(BOW) or term-frequency inverse-document-frequency~(TF-IDF) as features and then calculates the inner-product or cosine similarity between the feature vectors of questions and answers. Although this category of methods are simple, they cannot achieve satisfactory performance by discarding the syntactic and semantic information. The third category treats each answer as a class label and represents each question by BOW or TF-IDF features, based on which the QA task is treated as a classification task. This category of methods are mainly for factoid QA tasks~\cite{Boyd-Graber:12}.

The traditional QA methods mentioned above are usually called shallow methods. Recently, researchers propose to adopt deep methods, especially deep neural network~(DNN), for QA tasks~\cite{Hermanny:15,Iyyer:14,Tan:15,Wang:15}. Different from traditional shallow methods which use manually constructed high-dimensional feature vectors to represent questions and answers, deep methods try to learn low-dimensional distributed representation for questions and answers~\cite{Hermanny:15,Iyyer:14,Tan:15,Wang:15}. Furthermore, many works~\cite{Iyyer:14,Tan:15} have shown that deep methods can achieve better performance than traditional shallow methods for QA. Hence, researchers have put more and more attention to the deep QA methods. The existing deep QA methods can be divided into two main categories. The first category is based on recursive neural network~\cite{Iyyer:14}, and the second category is based on recurrent neural network~(RNN)~\footnote{In some literatures, the recursive neural network is also abbreviated as RNN. In this paper, we directly use the full name of recursive neural network, and RNN only represents the recurrent neural network.}~\cite{Hermanny:15,Tan:15,Wang:15}.

QANTA~\cite{Iyyer:14} is one of the representative deep QA methods based on recursive neural network. It is mainly developed for factoid QA. The inputs of QANTA are dependency parse trees of sentences in a question and its corresponding answer.  After learning the feature representations for the questions, QANTA trains a logistic regression~(LR) classifier on these feature representations and then uses it to classify the questions into their corresponding answers. Experiments show that QANTA outperforms traditional shallow methods. The disadvantage of recursive neural network based methods, such as QANTA, is that they need to build parse trees before the deep representation learning procedure .

Traditional RNN is unidirectional. In many NLP applications, especially those with long sequences, researchers find that bidirectional RNN~(BRNN) can outperform unidirectional RNN. Hence, BRNN has been widely used for QA tasks~\cite{Tan:15,Hermanny:15,Wang:15} with promising performance. \mbox{QA-LSTM}~\cite{Tan:15} first uses a gated BRNN, called bidirectional long-short term memory~(BLSTM), to extract the feature of questions (answers) and concatenates the hidden states at time step $t$ of both directions in BLSTM to generate the output at time step $t$. Then a pooling operation is performed on these outputs to generate the representations for questions~(answers). Loss or similarity is calculated after this pooling operation. \cite{Hermanny:15} uses an attention mechanism on the hidden states to generate the representations for questions. \cite{Wang:15} concatenates the question and answer as one sequence, then uses an output layer to compute similarity directly. A pooling operation is also applied to the outputs during training.  Different from recursive neural network based methods like QANTA which need some manual effort for building the parse trees, BRNN based deep QA methods can be used to train feature representation automatically in an end-to-end way. Furthermore, some work~\cite{Li:15} has shown that BRNN can outperform recursive neural network based methods in many NLP tasks including QA.

Although BRNN has achieved promising performance for QA, most existing BRNN models, such as QA-LSTM, extract the information of questions and answers by directly using a pooling operation to generate the representation for loss or similarity calculation. Hence, these existing models don't put supervision~(loss or similarity calculation) at every time step, which will lose some useful information. In this paper, we propose a novel BRNN model called full-time supervision based \mbox{BRNN}~(FTS-BRNN), for QA. The contributions of FTS-BRNN are briefly outlined as follows:
\begin{itemize}
\item FTS-BRNN can put supervision at every time step to make full use of all information in all time steps in BRNN.
\item Different from existing BRNN methods which use LSTM as the hidden units, FTS-BRNN uses the gated recurrent unit~(GRU)~\cite{Cho:15} as the hidden units.
\item Experiments on the factoid QA task show that FTS-BRNN can outperform other baselines to achieve the state-of-the-art accuracy in real applications.
\end{itemize}

The remaining content of this paper is organized as follows: Section~\ref{sec:background} introduces the background of this paper, including a short overview of factoid QA, BRNN and GRU; Section~\ref{sec:model} describes the model details of FTS-BRNN; Section~\ref{sec:experiment} presents the experimental results, and we draw our conclusion in Section~\ref{sec:conclusion}.

\section{Background}\label{sec:background}
Although our model can also be generalized to handle general QA tasks, this paper focuses on a special QA task, called factoid QA. In this section, we introduce the background of this paper, including a short overview of factoid QA, BRNN and GRU.

\subsection{Factoid Question Answering}

Factoid QA is a special QA task in which the answers are syntactic or semantic entities, such as organization or person names. Quiz bowl is a representative kind of factoid QA. Here, we use quiz bowl as an example to introduce factoid QA~\cite{Iyyer:14}. Quiz bowl can be seen as a kind of text classification task~\cite{Boyd-Graber:12} or QA task~\cite{Iyyer:14}. Players are asked for an answer according to the given description~(question). Each question consists of four to six sentences in which every single sentence contains useful clues to answer the question. The answer to a question is an entity represented by a phrase or a single word. Table~\ref{quizbowl} shows an example of quiz bowl. In real world competition, players can answer at any time. In our experiments, we slightly change the quiz bowl rule that players can only answer the question after getting all of the question information, which can better reflect the information extraction ability of the models and make the trained model be suitable for more general QA tasks.

\begin{table}[!htbp]
\caption{An example of quiz bowl.}
\label{quizbowl}
\centering
\small
\begin{tabular}{ll}
\toprule
         & In one novel set in this country, boats laden with straw dummies feign an attack\\
         & to steal enemy arrows for reuse.\\
         \cmidrule{2-2}
         & Another novel set in this country features a talking stone in its preface.\\
         \cmidrule{2-2}
         & The Oath of the Peach Garden occurs in one novel set in this country, which was\\
Question & also the setting of a novel in which one hundred and eight outlaws stow away in\\
         & a marsh.\\
         \cmidrule{2-2}
         & A Buddhist monk's travels with the Monkey King make up another of its "Four \\
         & Classical Novels".\\
         \cmidrule{2-2}
         & For 10 points, name this country, the setting of Water Margin, Journey to the\\
         & West, and Romance of the Three Kingdoms.\\
\midrule
Answer   & China\\
\bottomrule
\end{tabular}
\end{table}

\subsection{Bidirectional Recurrent Neural Network~(BRNN)}

RNN is proposed for processing sequential data containing several time steps, which has been widely used in NLP tasks, including neural machine translation~(NMT)~\cite{Sutskever:14,Bahdanau:15} and QA and so on. Traditional RNN is unidirectional. BRNN consists of two unidirectional RNNs in opposite directions, a forward RNN and a backward RNN. So the hidden states $h^{(t)}$ of BRNN at time step $t$ consist of the hidden states of the forward RNN $f^{(t)}\in\mathbb{R}^d$ and the hidden states of the backward RNN $b^{(t)}\in\mathbb{R}^d$. There are several ways to combine these two hidden states. In~\cite{Bahdanau:15,Tan:15}, a concatenating operation $h^{(t)} = [f^{(t)} , b^{(t)}]$ is adopted. In~\cite{Li:15}, an output activation function $o^{(t)} = f(W_L \cdot [f^{(t)}, b^{(t)}])$ is used, which can preserve the vector dimensionality with $W_L \in \mathbb{R}^{d \times 2d}$ where $d$ is the dimensionality of $f^{(t)}$ and $b^{(t)}$. The way used in~\cite{Wang:15} is similar to that in~\cite{Li:15}, which uses $ o^{(t)} = W \cdot f^{(t)} + U \cdot b^{(t)} + bias$ without activation functions. BRNN has achieved better performance than unidirectional RNN in real applications, especially with long sequences.
%

\subsection{Gated Recurrent Unit~(GRU)}
In real applications, the gated RNN or gated BRNN is always adopted. Typical gated RNN includes long short-term memory~(LSTM)~\cite{Hochreiter:97} and gated recurrent unit~(GRU). GRU is first proposed in~\cite{Cho:15} for NMT task. There are three gates in a single LSTM unit: input gate $i$, forget gate $f$ and output gate $o$. But GRU only uses two gates, reset gate $r$ and update gate $z$, to achieve similar functionality as that in LSTM.

In this paper, we choose GRU rather than LSTM because we find that in our model GRU is better than LSTM. Here, we give a brief introduction of GRU.

In GRU, reset gate $r$ and update gate $z$ are defined as follows:
\begin{gather}
r = \sigma(W_r x + U_r h^{(t-1)} + b_r), \nonumber \\
z = \sigma(W_z x + U_z h^{(t-1)} + b_z), \nonumber
\end{gather}
where $W_r, W_z, U_r, U_z \in\mathbb{R}^{d \times d}$ are weight matrices, $b_r, b_z$ are the bias vectors, $x$ is input and $h^{(t-1)}$ is the hidden states at the previous time step, $\sigma$ is the sigmoid function. Then the hidden states at time step $t$ are computed by
\begin{equation}\label{eq:GRUhiddenStates}
h^{(t)} = z \odot h^{(t-1)} + (1-z) \odot \widetilde{h}^{(t)},
\end{equation}
where
\begin{equation}
\widetilde{h}^{(t)} = \phi(W_h x + U_h(r \odot h^{(t-1})) + b_h), \nonumber
\end{equation}
with $W_h, U_h \in\mathbb{R}^{d \times d}$ being the weight matrices and $b_h$ being the bias vector.

\section{Full-Time Supervision based BRNN~(FTS-BRNN)}\label{sec:model}

In this section, we present the details of our full-time supervision based BRNN~(FTS-BRNN) which puts supervision for all time steps in BRNN.

FTS-BRNN has two variants: the first one adopts BRNN for questions and RNN for answers, and the second one adopts the same BRNN for both questions and answers. In this paper, FTS-BRNN refers to the first variant, and FTS-BRNN-s refers to the second variant which uses the \underline{s}ame BRNN for both questions and answers. BRNN is typically better than RNN for long sequences, but for short sequences BRNN is not necessarily better than RNN. Hence, if the answers are short, such as the case of factoid QA, we prefer to choose FTS-BRNN which adopts RNN for answers. This will also be verified in our experiments.

\subsection{FTS-BRNN}

\begin{figure}[t]
  \centering
  \includegraphics[width=1\textwidth]{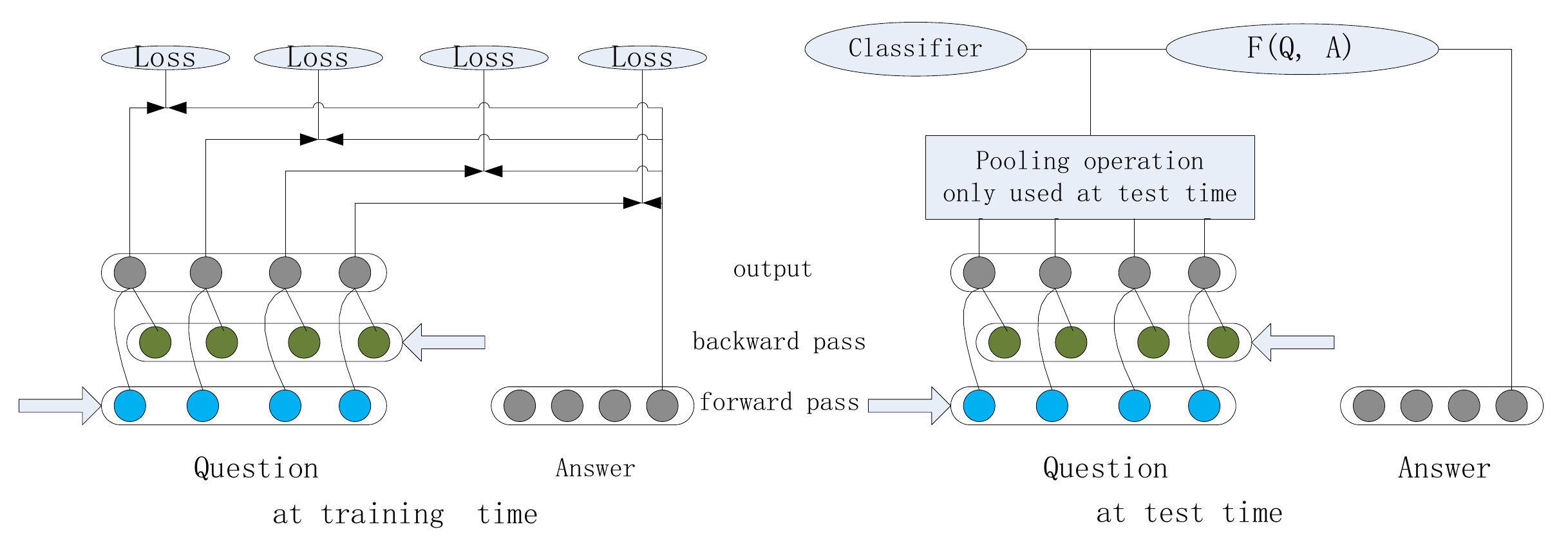}
  \caption{\label{model1}The architecture of FTS-BRNN.}
\end{figure}

The architecture of FTS-BRNN is shown in Figure~\ref{model1}. FTS-BRNN uses a BRNN for questions and a RNN for answers. The unit to represent the hidden states in each time step is a GRU. Hence, the BRNN in FTS-BRNN is actually a bidirectional GRU, containing a forward GRU and a backward GRU. Furthermore, different from some existing BRNN methods like QA-LSTM~\cite{Tan:15} which directly concatenate the hidden states of the forward RNN and backward RNN as output, FTS-BRNN adds an output layer on BRNN. The output of the time step $t$ is computed as follows:
\begin{equation}
o^{(t)} = W_o \cdot f^{(t)} + U_o \cdot b^{(t)} + bias, \nonumber
\end{equation}
where $f^{(t)} \in \mathbb{R}^d$ is the hidden states of the time step $t$ in the forward GRU, $b^{(t)} \in \mathbb{R}^d$ is the hidden states of the time step $t$ in the backward GRU, $bias \in \mathbb{R}^d$ is a bias vector, $W_o,U_o \in \mathbb{R}^{d \times d}$ are weight matrices. Here, all the hidden states of both forward GRU and backward GRU are computed by~(\ref{eq:GRUhiddenStates}).

The model of FTS-BRNN can be formulated as follows:
\begin{align}
Q_x^f &= BRNN_f(x), \nonumber \\
Q_x^b &= BRNN_b(x), \nonumber \\
Q_x^o &= W_o \cdot Q_x^f  +  U_o \cdot Q_x^b + bias, \nonumber \\
A_x^h &= RNN(A_x), \nonumber \\
A_x^e &= A_x^h{(T_a)}, \nonumber
\end{align}
where $x$ is a question, $Q_x^f = [f_x^{(1)}; f_x^{(2)}; ...;f_x^{(T_q)}] \in \mathbb{R}^{d \times T_q}$ is the forward hidden states at all the $T_q$ time steps, $Q_x^b = [b_x^{(1)}; b_x^{(2)}; ...;b_x^{(T_q)}] \in \mathbb{R}^{d \times T_q},Q_x^o= [o_x^{(1)}; o_x^{(2)}; ...;o_x^{(T_q)}] \in \mathbb{R}^{d \times T_q}$ are the backward hidden states and outputs respectively, $T_q$ and $T_a$ are respectively the length of the question and answer, $A_x$ is the answer of $x$, $A_x^h \in \mathbb{R}^{d \times T_a}$ are the hidden states of the answer at all the $T_a$ time steps, $A^e_x \in \mathbb{R}^{d}$ is the hidden states of the last time step $T_a$ of the RNN.

During the training procedure, FTS-BRNN minimizes the full-time margin loss:
\begin{equation}\label{eq:fulltimeLoss}
Loss = \sum_x \sum_t max(0, 1 - o_x^{(t)} \cdot A^e_x + o_x^{(t)} \cdot A^e_{wrong})
\end{equation}
where $A^e_x$ is the corresponding answer of the question $x$ and $A^e_{wrong}$ is a wrong answer for question $x$. We use all the wrong answers in our experiments rather than just randomly sample a subset to calculate the loss. This margin loss aims to make the inner product between the question representation and the corresponding answer representation bigger than those of the wrong answers as much as possible.

Most existing methods, such as QA-LSTM~\cite{Tan:15}, use pooling to generate the question representation. Based on the pooling result, the loss of these methods has the following formulations:
\begin{equation}\label{eq:poolingLoss}
Loss = \sum_x  max(0, 1 - o^p_x \cdot A^e_x + o_x^p \cdot A^e_{wrong}),
\end{equation}
where $o^p_x$ is the question representation after pooling. For example, $o^p_x = \frac{\sum_t o_x^{(t)}}{T_q}$ is the result of average pooling. We can also use other pooling operations to generate $o^p_x$, and we can also use other loss functions besides the margin loss.

By comparing the loss in~(\ref{eq:fulltimeLoss}) to that in~(\ref{eq:poolingLoss}), we can find that FTS-BRNN puts supervision for all the time steps. This full-time supervision strategy in FTS-BRNN can make better use of the information. Since each output $o_x^{(t)}$ has all the information of input in BRNN, full-time supervision treats these outputs as representations of questions independently. By minimizing this full-time supervised loss, every $o_x^{(t)}$ tries to make the inner product between $o_x^{(t)}$ and $A^e_x$ bigger than the inner product between $o_x^{(t)}$ and $A^e_{wrong}$.

After training, the distributed representation of a question is computed by an average pooling operation for out-of-sample prediction~(test):
\begin{equation}
Q_k^p = \frac{\sum_t o_k^{(t)}}{T_q}
\end{equation}
where $k$ is a question for test~(prediction), $Q_k^p$ is the representation for question $k$ after average pooling.

Then, the answer which gives the biggest inner product with $Q_k^p$ will be chosen:
\begin{equation}\label{eq:innerProductPrediction}
y = \mathop{\argmax}_i (Q_k^p \cdot A^e_i).
\end{equation}

Since the output $o_x^{(t)}$ at each time step contributes to the loss in FTS-BRNN during training, the prediction function in~(\ref{eq:innerProductPrediction}) plays a role like ensemble by using the average pooling operation on the test questions.

After we have learned the representation for all questions, we can also treat each answer as a class label, and then train a logistic regression~(LR) classifier on question representations to predict the answer:
\begin{equation}\label{eq:LRPrediction}
y = LR(Q_k^p).
\end{equation}


In~\cite{Li:15}, the authors also use similar loss as that in~(\ref{eq:fulltimeLoss}). However, the motivation of~\cite{Li:15} is to perform fair comparison with QANTA because QANTA also uses loss~(supervision) at each node~(step) in the recursive neural networks. Hence, the authors of~\cite{Li:15} do not explicitly claim that full-time supervision is the key in BRNN for QA tasks because they do not perform any empirical comparison between full-time supervision and pooling-based supervision. In this paper, we perform detailed empirical comparison between full-time supervision and pooling-based supervision, and find that full-time supervision is much better than pooling-based supervision. Hence, our work is the first to explicitly claim that full-time supervision is the key in BRNN for QA. Furthermore, LSTM is adopted in~\cite{Li:15}, but our FTS-BRNN adopts GRU. From our experiments which will be presented below, we find that FTS-BRNN with GRU is much better than FTS-BRNN with LSTM.

\subsection{FTS-BRNN-s}
\begin{figure}[t]
  \centering
  \includegraphics[width=1\textwidth]{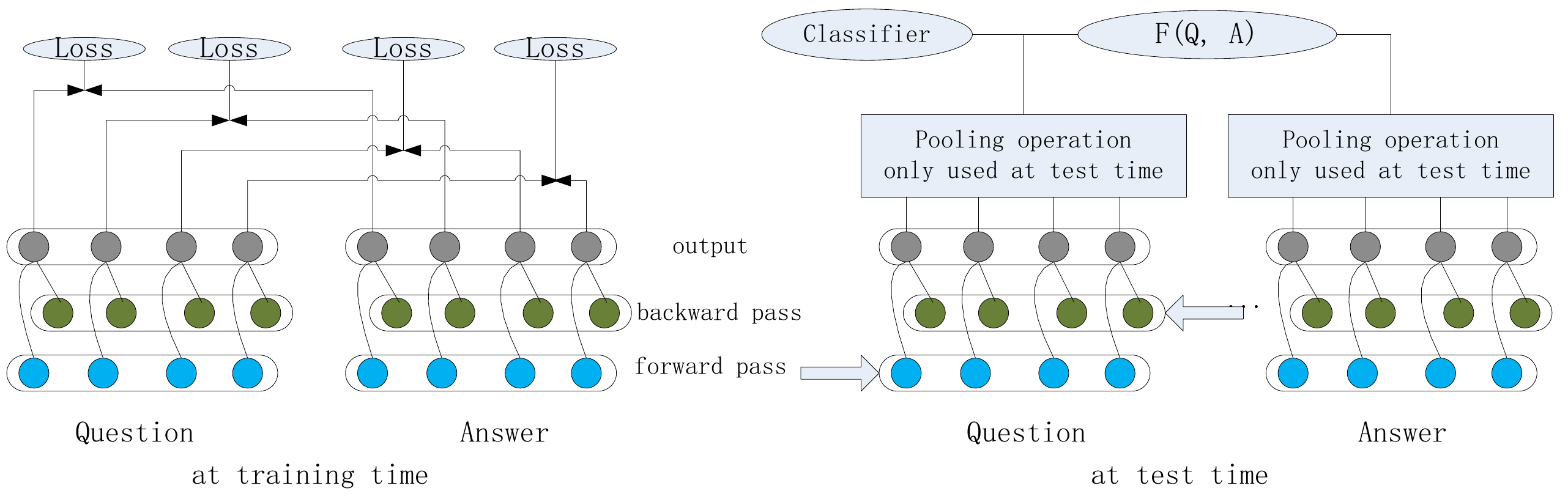}
  \caption{\label{model2}The architecture of FTS-BRNN-s.}
\end{figure}

Here, we introduce a variant of FTS-BRNN, which is called FTS-BRNN-s. Here, the `s' means that we use the \underline{s}ame BRNN for both questions and answers. The architecture of FTS-BRNN-s is shown in Figure~\ref{model2}. The question processing in FTS-BRNN-s is the same as that in FTS-BRNN. The only difference lies in the processing of answers. Different from $A^e_x$ of FTS-BRNN, in FTS-BRNN-s we get $A^o_x$ as follows:
\begin{align}
A^f_x&= BRNN_f(A_x), \nonumber \\
A^b_x &= BRNN_b(A_x), \nonumber\\
A^o_x &= W_o \cdot A^f_x  + U_o \cdot A^b_x + bias.\nonumber
\end{align}
Then the full-time margin loss is changed to:
\begin{equation}
Loss = \sum_x \sum_t max(0, 1 - o_x^{(t)} \cdot A_x^o{(t)} + o_x^{(t)} \cdot A^o_{wrong}{(t)}), \nonumber
\end{equation}
where $A_x^o{(t)}$ denotes the output at the time step $t$ for the answer of question $x$, $A^o_{wrong}{(t)}$ denotes the output at the time step $t$ for a wrong answer of question $x$.

At test~(prediction) time, the distributed representation of the answer $k$ is computed by an average pooling operation:
\begin{equation}
A^p_k = \frac{\sum_t A_k^o{(t)}}{T}, \nonumber
\end{equation}
where $T = T_a = T_q$ is the defined length of sequence.

\section{Experiments}\label{sec:experiment}

We evaluate our method on the factoid QA task. The experiments are performed on an NVIDIA K80 GPU server.

\subsection{Dataset}

We use the factoid QA datasets from~\cite{Iyyer:14} for evaluation\footnote{We download the datasets from \url{https://cs.umd.edu/~miyyer/qblearn/} which are provided by the authors of~\cite{Iyyer:14}. The publically available datasets for download are slightly smaller than those used in~\cite{Iyyer:14}.}. The whole dataset contains two subsets: Literature and History. Similar to~\cite{Iyyer:14}, we first filter out all questions that do not belong to history or literature, and then only the answers that occur at least six times will be used. The statistics of these two subsets are summarized in Table~\ref{dataset}.
\begin{table}[h]
\caption{The datasets. }
\label{dataset}
\centering
\begin{tabular}{lll}
\toprule
\bf subset & \bf \#questions & \bf \#answers \\
\midrule
Literature & 4204 & 424 \\
History & 2557 & 303 \\
\bottomrule
\end{tabular}
\end{table}

For all questions belonging to the same answer, we sample 20\% as test set, 20\% as validation set, and the remaining 60\% as training set. So we get 2524 training questions, 840 validation questions and 840 test questions for Literature. For History, we have 1535 training questions, 511 validation questions and 511 test questions.

\subsection{Baselines}\label{sec:baseline}
We compare our method with several state-of-the-art baselines:
\begin{itemize}
\item {\emph{BOW}}~\cite{Iyyer:14}: BOW treats each answer as a class label, and adopts the LR classifier for classification based on the bag-of-words~(BOW) features.

\item {\emph{BOW-DT}}~\cite{Iyyer:14}: BOW-DT is similar to BOW method by using LR on BOW features. Different from BOW, the feature set is augmented with dependency relation indicators.

\item {\emph{QANTA}}~\cite{Iyyer:14}: QANTA is a recursive neural network based method proposed in~\cite{Iyyer:14}. During the training procedure, QANTA adopts dependency parse trees to learn sentence representation and then trains a LR classifier based on this representation. During prediction~(test), the answer is chosen by the LR classifier.

\item {\emph{QA-LSTM}}~\cite{Tan:15}: QA-LSTM is a LSTM based BRNN method~\cite{Tan:15} which computes the loss after using a pooling operation and uses concatenating operation to generate the hidden states $h^{(t)}$ for each time step.
\end{itemize}



Because the question is a paragraph consisting of several sentences, we concatenate these sentences one by one as a single sentence\footnote{We also try the hierarchical RNN~\cite{LiHierarchical:15,Lin:15} which aims to deal with paragraphs or documents. But it does not bring us better performance.}. As that in QANTA~\cite{Iyyer:14}, we use the 100-dimensional pre-trained word embedding provided by GloVe~\cite{Pennington:14} to represent the input words for all deep methods. Furthermore, we set the dimensionality of hidden states and embedding $d = 100$ to keep in line with QANTA. We choose rmsprop and momentum as our training algorithm. Learning rate is 0.002 or 0.001 for different methods to achieve the best performance, and the momentum is 0.8. Dropout is performed on the inputs of questions with rate 0.7. All methods are converged around 50-100 epochs. The weight matrices and initial states of RNN are initialized by a uniform distribution $[-a, a]$, where $a = \sqrt{\frac{6}{InputSize + OutputSize}}$ is related to the size of input and output.


\subsection{Results}
We first perform experiments to verify the effectiveness of the full-time supervision strategy and the output layer in FTS-BRNN, and then verify the advantage of GRU against LSTM. Finally, we compare our method with other state-of-the-art baselines to show the promising performance of FTS-BRNN. All the results are based on the metric of prediction accuracy. "InnerP" represents the results with inner product for prediction as shown in~(\ref{eq:innerProductPrediction}), and "LR" represents the results with LR for prediction as shown in~(\ref{eq:LRPrediction}).

\subsubsection{Effect of Full-Time Supervision and Output Layer}
Because FTS-BRNN always needs the output layer to make the dimensionality of answers and questions be equal, we demonstrate the effect of full-time supervision and output layer based on FTS-BRNN-s.  Table~\ref{tab:result3} shows the results with different configurations for FTS-BRNN and \mbox{FTS-BRNN-s}. Here, "Pooling-loss" is the loss used in~\cite{Tan:15} and (\ref{eq:poolingLoss}) while "FTS-loss" is our loss function in~(\ref{eq:fulltimeLoss}) for full-time supervision. "has-output" represents the model with an output layer $o^{(t)} = W_o \cdot f^{(t)} + U_o \cdot b^{(t)} + bias$ while "no-output" represents the model without an output layer by directly using concatenating operation to get the hidden states $h^{(t)} = [f^{(t)} , b^{(t)}]$. The only difference between "FTS-BRNN-s with pooling loss" and "FTS-BRNN-s" is that "FTS-BRNN-s with pooling loss" adopts the loss in~(\ref{eq:poolingLoss}) and ``FTS-BRNN-s" adopts the loss in~(\ref{eq:fulltimeLoss}). We can find that the FTS-BRNN-s with full-time supervision can dramatically outperform the counterpart with pooling loss, which successfully verifies the effectiveness of full-time supervision. Furthermore, we can also find that the output layer in FTS-BRNN and FTS-BRNN-s is also very important. In addition, FTS-BRNN is slightly better than FTS-BRNN-s for this factoid QA task.
\begin{table}[htbp]
\caption{Effect of full-time supervision and output layer.}
\label{tab:result3}
\centering
\begin{tabular}{cccccc}
\toprule
Model          &  Configuration                 & \multicolumn{2}{c}{Literature} & \multicolumn{2}{c}{History}\\
\midrule
               &                         & InnerP & LR & InnerP & LR      \\
\midrule
FTS-BRNN-s with pooling loss    & Pooling-loss, has-output & 77.0 & 83.5 & 70.0 & 80.6   \\
FTS-BRNN-s without output & FTS-loss, no-output & 28.7 & 85.5 & 41.1 & 74.4 \\
FTS-BRNN-s   & FTS-loss, has-output     & 89.2 & 93.0 & 82.4 & 87.9     \\
FTS-BRNN     & FTS-loss, has-output     & 89.8 & 93.1 & 83.6 & 88.1 \\
\bottomrule
\end{tabular}
\end{table}

\subsubsection{Effect of GRU}

The accuracy comparison between GRU and LSTM is shown in Table~\ref{tab:result2}, where "FTS-BRNN" is the method proposed in this paper with GRU for BRNN and "FTS-BRNN with LSTM" denotes a variant by substituting GRU with LSTM. We can find that GRU can outperfom LSTM in our FTS-BRNN model. Hence, our FTS-BRNN adopts GRU for BRNN.

\begin{table}[htbp]
\caption{Comparison between GRU and LSTM.}
\label{tab:result2}
\centering
\begin{tabular}{ccccc}
\toprule
             &  \multicolumn{2}{c}{Literature} & \multicolumn{2}{c}{History} \\
\midrule
Model               & InnerP  & LR  & InnerP  & LR  \\
\midrule
FTS-BRNN with LSTM  & 86.2   & 89.5  & 79.1  & 81.8 \\
FTS-BRNN   & 89.8   & 93.1  & 83.6  & 88.1\\
\bottomrule
\end{tabular}
\end{table}
\subsubsection{Comparison to Baselines}
Table~\ref{tab:result1} reports the accuracy comparison between our method and other state-of-the-art baselines introduced in Section~\ref{sec:baseline}.  Because the public scripts of BOW, BOW-DT and QANTA don't use inner product to choose answers, we don't report inner product results of these three methods.

\begin{table}[htbp]
\caption{Accuracy comparison to other baselines.}
\label{tab:result1}
\centering
\begin{tabular}{ccccc}
\toprule
             &  \multicolumn{2}{c}{Literature} & \multicolumn{2}{c}{History} \\
\midrule
Model               & InnerP   & LR  & InnerP   & LR  \\
\midrule
BOW                 &  ---    & 46.2 & ---   & 50.8\\
BOW-DT              &  ---    & 57.4 & ---   & 60.9\\
QANTA               &  ---    & 63.0 & ---   & 65.8\\
QA-LSTM             & 78.7   & 86.9        & 69.7   & 80.2 \\
\midrule
FTS-BRNN-s& 89.2   & 93.0        & 82.4   & 87.9\\
FTS-BRNN   & 89.8   & 93.1        & 83.6   & 88.1\\
\bottomrule
\end{tabular}
\end{table}

From Table~\ref{tab:result1}, we can find that the results of LR are better than those of inner product. All the deep methods, including QANTA, QA-LSTM, FTS-BRNN-s and FTS-BRNN, can outperform traditional non-deep methods. Furthermore, all the RNN-based methods, including QA-LSTM, FTS-BRNN-s and FTS-BRNN, can outperform recursive neural network based method~(QANTA). In addition, our FTS-BRNN and FTS-BRNN-s can outperform all the other state-of-the-art baselines to achieve the best performance.

\section{Conclusion}\label{sec:conclusion}
In this paper, we have proposed a full-time supervision based bidirectional RNN method, called FTS-BRNN, for QA tasks. This is the first work to perform detailed empirical comparison between full-time supervision and pooling-based supervision and explicitly claim that full-time supervision is the key in BRNN for QA. Furthermore, we also find that GRU is better than LSTM for BRNN based QA. Experiments on factoid QA task show that our FTS-BRNN method can outperform other state-of-the-art baselines in real applications. In our future work, we will apply our method to other QA tasks, especially those with long answers.

\bibliographystyle{plain}
\bibliography{FTS-BRNN_arxiv}
\end{document}